\documentclass{article}
\usepackage[T1]{fontenc}
\usepackage{iclr2027_conference,times}
\usepackage{amsmath,amsfonts,bm}

\usepackage{url}
\usepackage{booktabs}
\usepackage{graphicx}
\usepackage{xcolor}
\usepackage{colortbl}
\usepackage{tcolorbox}
\usepackage{capt-of}
\usepackage{placeins}
\usepackage{wrapfig}
\usepackage{fontawesome5}
\usepackage[normalem]{ulem}

\definecolor{findingborder}{HTML}{205D60}
\definecolor{tablegroup}{HTML}{E8F0EF}
\newcommand{\finding}[2]{%
  \begin{tcolorbox}[colback=black!3,colframe=findingborder,boxrule=0.45pt,
    arc=2pt,left=7pt,right=7pt,top=6pt,bottom=6pt,
    before skip=5pt,after skip=7pt]
    \textbf{Finding #1:} #2
  \end{tcolorbox}}
\usepackage{amssymb}
\usepackage{hyperref}
\usepackage{cleveref}

\definecolor{rosegold}{rgb}{0.72, 0.43, 0.47}

\Crefname{section}{Section}{Sections}
\Crefname{table}{Table}{Tables}
\crefname{section}{Sec.}{Secs.}
\crefname{table}{Tab.}{Tabs.}
\crefname{figure}{Fig.}{Figs.}
\crefname{appendix}{Sec.}{Secs.}
\crefname{equation}{Eq.}{Eqs.}

\hypersetup{
  pdftitle={MetaRubric: Learning to Reward for Rubric-Based Reinforcement Learning},
  pdfauthor={Yuxuan Fan and Jaehong Yoon},
  colorlinks=true,
  citecolor=rosegold,
  linkcolor=blue,
  urlcolor=blue
}

\title{MetaRubric: Learning to Reward for\\
Rubric-Based Reinforcement Learning}

\author{%
\begin{tabular}{cc}
  \makebox[0.4\textwidth][c]{\textbf{Yuxuan Fan}} &
  \makebox[0.4\textwidth][c]{\textbf{Jaehong Yoon}\thanks{Corresponding author.}} \\
  NTU Singapore & NTU Singapore
\end{tabular}
\\[5ex]
  \href{https://metarubric.github.io}{\faGlobe\ \texttt{https://metarubric.github.io}}}

\makeatletter
\renewcommand{\@maketitle}{%
  \vbox{\hsize\textwidth
    {\LARGE\scshape\@title\par}
    \vskip 1.5em
    {\centering\normalfont\@author\par}
    \vskip 0.3in minus 0.1in}}
\makeatother

\newcommand{\method}{MetaRubric}

\iclrfinalcopy

\begin{document}

\setcounter{footnote}{1}
\maketitle
\lhead{Preprint.}

\begin{abstract}
Rubric-based reinforcement learning extends reward-driven optimization to open-ended tasks by assigning partial credit to individual response requirements.
However, rubric judges can assign a high criterion score even when the information or action it requires is absent from the response, a failure mode we term \textit{Vacuous Credit}.
Such awards persist after the required information is removed and can reverse the sign of a response's GRPO advantage.
To address this problem, we introduce MetaRubric, which alternates evidence-aware policy optimization with response-guided rubric adaptation.
We construct counterfactual counterparts by changing one task-relevant fact in each prompt.
During policy optimization, credit is assigned only when the response contains sufficient evidence to satisfy the required rubric criterion.
After each policy-optimization stage, current policy responses guide revisions to original and counterfactual criteria while preserving the meaning of the original prompt's initial rubric as interpreted under each prompt's facts.
We also adapt criterion weights at stage boundaries to better address observed policy errors.
Across multiple backbones, MetaRubric improves PubMedQA accuracy by 6.00--20.40 percentage points over static-judge GRPO, with further gains on HealthBench-Hard and two multimodal medical benchmarks.
\end{abstract}

\section{Introduction}
\label{sec:introduction}

Rubric-based reinforcement learning (RL) provides fine-grained rewards for open-ended tasks by assigning partial credit across instance-specific response criteria~\citep{liu2026openrubrics,shan2026survey}.
A language-model judge evaluates these criteria, and their weighted scores provide the reward used to optimize the policy, even for tasks with multiple valid responses and no single automatically verifiable answer~\citep{gunjal2025rar,li2026rubrichub,ye2025selfrewarding}.
A critical challenge arises when the reward credits information or actions absent from the response, so that policy optimization favors responses that fail important requirements~\citep{mahmoud2026rubrichacking}.
\begin{figure}[t]
    \centering
    \includegraphics[width=\linewidth]{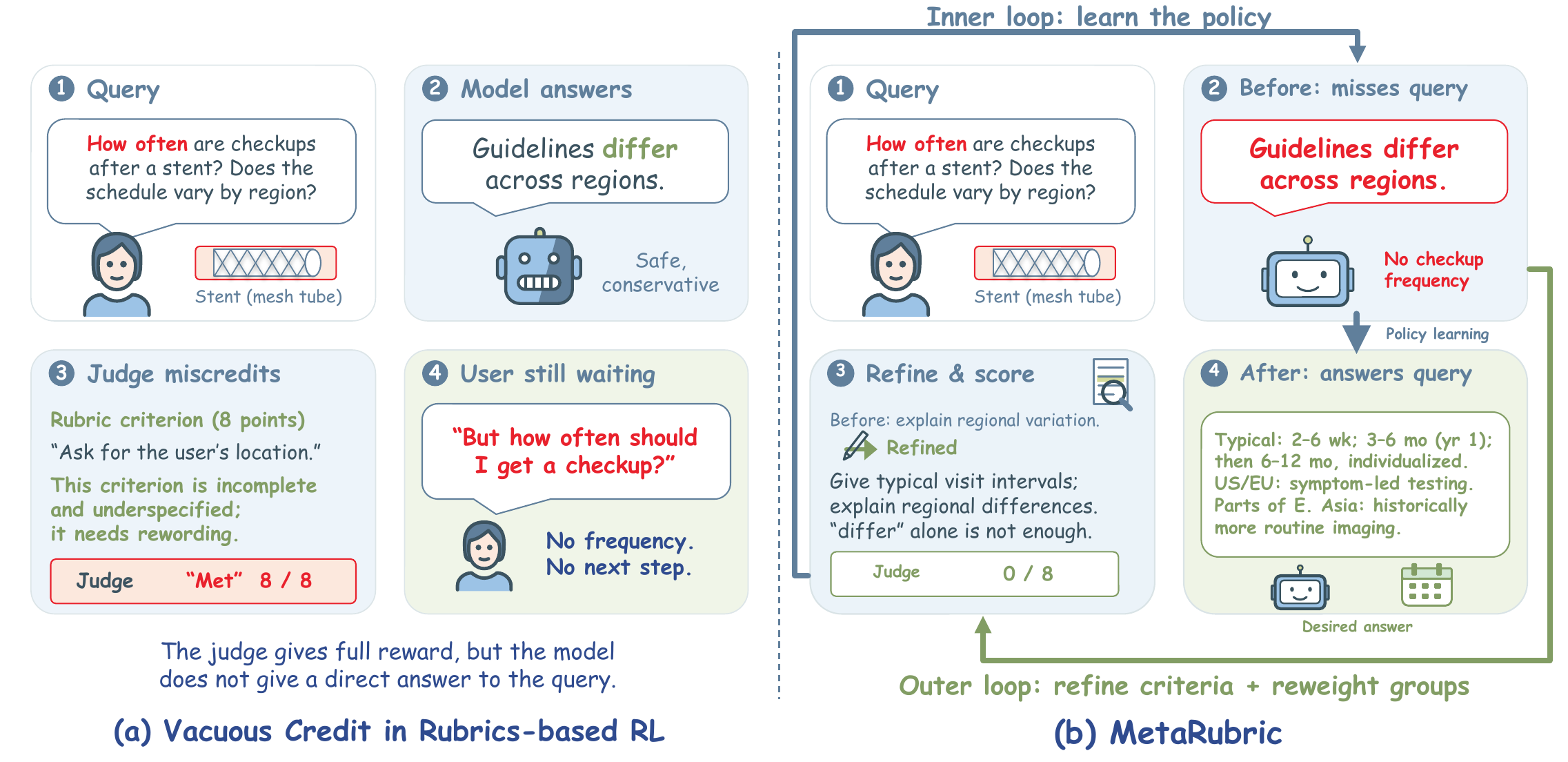}    \vspace{-0.25in}
    \caption{\textbf{Motivation and overview of MetaRubric.}
(a) \textit{Vacuous Credit} in rubric-based RL. Under an incomplete, underspecified criterion, the judge awards full credit to a safe, conservative response that leaves the query unanswered.
(b) 
MetaRubric addresses this by jointly improving the policy and rubric: the inner loop optimizes evidence-aware responses, while the outer loop reweights criterion groups and revises criterion descriptions.This produces clearer criteria and more concrete answers.
}
    \label{fig:intro-motivation}
\end{figure}

\cref{fig:intro-motivation}(a) illustrates a failure mode observed in our rubric-judge evaluations. Consider a patient asking how often to attend checkups after a stent procedure and whether the schedule varies by region. The rubric requires the model to ask for the patient's location, yet the model only notes that guidelines differ across regions and provides no checkup frequency. The judge nevertheless marks the location criterion as satisfied, awarding credit for an action the model never performed.
This poses two problems in rubric-based judging. (1) Criterion underspecification: the rubric may not clearly define the task-specific requirements or the observable evidence needed for credit. As a result, awards can persist even after all information required by a criterion is removed from the response (\cref{sec:vac-omission}). (2) Unsupported judging: the judge may award credit for information or actions absent from the response, distorting the relative rewards used for policy optimization (\cref{sec:vac-optimization}).

We call this broader failure \textit{Vacuous Credit}: a judge assigns a high criterion score even though the response omits information or an action required to satisfy it.
Such erroneous credit corrupts the learning signal: within a group relative policy optimization (GRPO) rollout group, correcting a single erroneous award can flip the sign of a response's advantage, reversing its contribution to the policy update. Repeated optimization can therefore favor responses that score well under the grading procedure without satisfying the intended requirement.

To address this challenge, we introduce \method{}, which alternates an inner loop of evidence-aware policy optimization with an outer loop of rubric adaptation (\cref{fig:intro-motivation}(b)).
Each prompt is paired with a counterfactual that changes one task-relevant fact, exposing which requirements depend on that fact.
In the inner loop, the policy is optimized under the current rubric. 
For each criterion, the judge assigns three scores: satisfaction, requirement coverage, and evidential support. Credit is bounded by the minimum of these three scores. The resulting rubric reward is combined with an auxiliary question-answering (QA) score to optimize the policy with GRPO.
At the end of each training stage, the outer loop 
updates the rubric in two steps. First, \textit{error-guided weight adaptation} adjusts criterion-group weights based on observed policy errors, subject to magnitude bounds and severity-order constraints. Second, semantically anchored \textit{rubric revision} updates criterion descriptions or correspondence labels using collected responses, while preserving the requirements of the initial rubric for each prompt.
Only revisions that improve agreement with held-out reference judgments and pass semantic review are accepted. 
The resulting weights and revisions define the rubric configurations for the next training stage.

We evaluate \method{} on text-only and multimodal medical benchmarks across multiple LLM and MLLM backbones. Across baselines, \method{} improves PubMedQA accuracy by 6.00--20.40 points, HealthBench-Hard accuracy by 2.46--3.19 points, MMOral-X by 1.29--2.13 points, and MMOral-OPG by 3.09--3.82 points.

\section{Related Work}
\label{sec:related}

\noindent\textbf{Rubric-based Reinforcement Learning.}
Rubrics as Rewards~\citep{gunjal2025rar} extends on-policy RL to open-ended tasks
using instance-specific criteria. RubricHub and OpenRubrics scale criterion
generation~\citep{li2026rubrichub,liu2026openrubrics,yuxuan2026phoenixnest}, while RRD refines their
granularity and weights~\citep{shen2026rrd}. Reward generation also uses pairwise
adaptive criteria~\citep{jia2026openrs}, policy self-grading~\citep{ye2025selfrewarding},
and joint optimization of rubric generators with
reasoners~\citep{sheng2026rlcer,guan2026evorubric} or
judges~\citep{xu2026rubricarm}. On the other hand, \method{} couples evidence-aware policy optimization with response-guided rubric adaptation. Policy errors drive criterion revision on original and counterfactual cases, as well as criterion-group reweighting, while preserving the initial rubric's intended requirements.

\noindent\textbf{Reward Hacking in Rubric-based RL.}
Proxy reward gains can diverge from response quality~\citep{gao2023scaling}.
Chasing the Tail~\citep{zhang2026chasingtail} improves rubric discrimination
among high-quality responses, while \citet{mahmoud2026rubrichacking} distinguish
verifier failures from rubric-design limitations. Mitigations include
deterministic verification with restricted input exposure in
RLR$^3$~\citep{yu2026rlr3} and random criterion dropout during
RL~\citep{yang2026rubricdropout}. We identify \emph{Vacuous Credit}, in which
a judge assigns a high criterion score to empty assurances, generic safety advice,
or defensive qualifications even when the required content is entirely absent.
Content-deletion tests and recorded-group
replay connect this judgment error to policy optimization: an award can survive
removal of its evidence, and removing an absent-content award can reverse a
GRPO advantage sign (\cref{sec:vacuous-credit}). \method{} makes
content coverage and support explicit components of credit assignment
and uses policy responses to adapt the rubric during training.

\section{Vacuous Credit in Rubric-Based Reinforcement Learning}
\label{sec:vacuous-credit}

We formulate rubric-based RL as policy optimization under criterion-level rewards.
Given a query prompt $x$ and its conversation context, we learn a policy $\pi_\theta(y|x)$ that generates responses $y$ satisfying a prompt-specific rubric.
Each criterion $k$ specifies a response requirement and carries an importance weight $w_k\geq0$.
A judge returns $m_k(x,y)\in\{0,1\}$, indicating whether the requirement is fulfilled.
The rubric assigns each criterion a binary reference value $b_k\in\{0,1\}$, which determines whether the source score is used directly or complemented on the fulfillment scale (\cref{app:scoring-details}).
We define $B_x=\sum_k w_k b_k$ as the reference offset for the rubric and $P_x=\sum_k w_k-B_x>0$ as its normalization factor.
The rubric reward is
\begin{equation}
 R(x,y)=\frac{\sum_k w_k m_k(x,y)-B_x}{P_x}.
 \label{eq:vac-main}
\end{equation}
Training maximizes $\mathbb E_{x\sim\mathcal D,\,y\sim\pi_\theta(\cdot\mid x)}[R(x,y)]$ over the training prompt distribution $\mathcal D$.
This objective assumes that higher judge rewards reflect better rubric fulfillment. We test this assumption empirically, examine whether criterion credit persists after removing required evidence, and analyze how erroneous credit distorts the GRPO learning signal.

\begin{figure}[t]
\centering
\includegraphics[width=\linewidth]{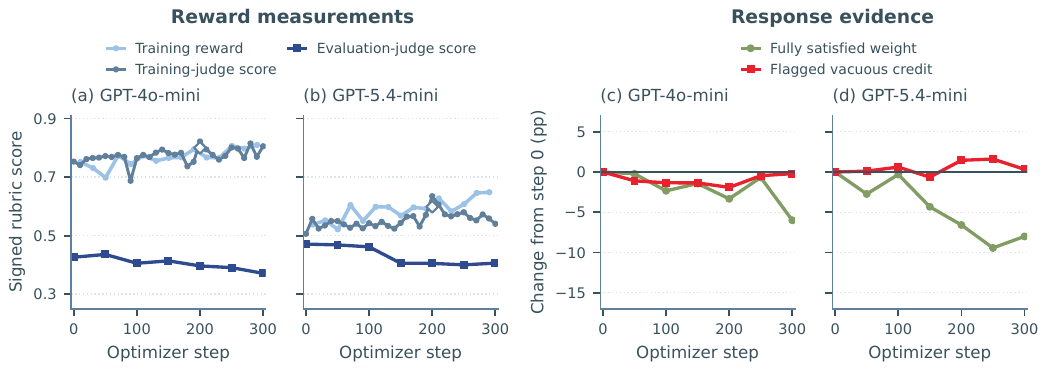}
\vspace{-0.3in}\caption{\textbf{Reward growth and rubric satisfaction.}
\textbf{(a,b)} Training reward and rubric scores during training; evaluation uses a panel of three stronger judges on a separate test set. Panels (a,c) use GPT-4o-mini as the training judge; panels (b,d) use GPT-5.4-mini.
\textbf{(c,d)} Changes from step 0 in fully satisfied requirement weight and Vacuous Credit identified using a stronger model to audit response evidence. Both are normalized by total available rubric credit $P_x$ and expressed in percentage points.}
\label{fig:vac-retraining}
\end{figure}

\subsection{Reward Growth Reveals Vacuous Credit}
\label{sec:vac-training}

\finding{1}{Training reward 
can rise even as
rubric satisfaction falls.}

Does a higher training reward always imply that the policy better satisfies the rubric? To test this, we train Qwen3-8B on HealthBench~\citep{arora2025healthbench} using GPT-4o-mini or GPT-5.4-mini as the reward judge. 
We evaluate the resulting policies on a held-out test set with a different prompt and three stronger judges: GPT-5.5, Gemini 3.5 Flash, and DeepSeek-V4-Pro. We count a criterion as satisfied only under unanimous agreement, and compute the final score using the rubric weights in \cref{eq:vac-main}.
Interestingly, as shown in \cref{fig:vac-retraining}(a,b), training reward increases even as evaluation-panel scores decline, revealing a clear mismatch between the optimized reward and rubric fulfillment.

\noindent\textbf{Vacuous Credit.}
To understand this gap, we inspect criterion-level judgments and find that a judge can assign a high criterion score even when the required information or action is entirely absent from the response.
As illustrated in \cref{fig:intro-motivation}(a), criterion credit often persists even when the required evidence is entirely missing from the response. We term this failure \textit{Vacuous Credit}.
Let $C_x$ denote the set of criteria with $b_k=0$ that require the response to provide specific information or perform an action. 
For each $k\in C_x$, we define $e_k(x,y)=1$ if the required information or action is completely omitted from $y$, and $0$ otherwise.
The reward attributable to Vacuous Credit is
\begin{equation}
 v(x,y)=\frac{1}{P_x}\sum_{k\in C_x}w_k m_k(x,y)e_k(x,y).
 \label{eq:vac-component}
\end{equation}
Subtracting $v(x,y)$ from $R(x,y)$ removes credit assigned to fully omitted requirements while leaving all other criterion contributions unchanged.

\noindent\textbf{Measuring rubric satisfaction and Vacuous Credit.}
We independently measure rubric satisfaction using stronger judges to audit the test responses, defined as the normalized weight of fully satisfied criteria.
\cref{fig:vac-retraining}(c,d) shows that rubric satisfaction fluctuates but declines overall as training reward increases, whereas auditor-flagged Vacuous Credit persists throughout training.
These trajectories reveal that omitted content can still receive credit during training. We next examine this failure through controlled evidence removal.

\subsection{Vacuous Credit Persists After Required Evidence Is Removed}
\label{sec:vac-omission}

\finding{2}{Most criterion awards 
persist after the required content is removed.
}

To directly verify whether criterion credit depends on the required evidence, we perform a deletion test.
Given 1,000 policy rollouts judged correct during training, 
we use GPT-5.5 to generate two paired edits of each response.
\textit{Required-content deletion} removes all information or actions needed to satisfy the target criterion while retaining generic statements
, whereas the control,
\textit{Generic-statement deletion} removes statements that do not fulfill the target requirement while preserving the required content.
We score the intact and edited responses using the reward judges and training reward prompt from \cref{sec:vac-training}. For each judge, award retention is the percentage of target criteria credited in the intact response that remain credited after deletion.

\begin{wraptable}{r}{0.36\textwidth}
\centering\small
\vspace{-0.35in}
\caption{\textbf{Retention rate} of target-criterion awards in the deletion test.}
\label{tab:vac-deletion}
\vspace{0.05in}
\setlength{\tabcolsep}{3pt}
\renewcommand{\arraystretch}{1.18}
\begin{tabular}{@{}l|cc@{}}
\toprule
& \multicolumn{2}{c}{Deleted content}\\
\cmidrule(l){2-3}
Reward judge & Required & Generic\\
& content & statements\\
\midrule
GPT-4o-mini & $82.0\%$ & $93.1\%$\\
GPT-5.4-mini & $69.8\%$ & $95.0\%$\\
\bottomrule
\end{tabular}
\vspace{-0.4in}
\end{wraptable}

As shown in \cref{tab:vac-deletion}, GPT-4o-mini and GPT-5.4-mini retain $82.0\%$ and $69.8\%$ of target-criterion awards after required-content deletion. Thus, removing required content lowers retention compared with the controlled setting, yet most awards persist even when the target requirement is no longer satisfied. These retained awards constitute \textit{Vacuous Credit} and directly affect the learning signal used for policy optimization.

\subsection{Implications for Policy Optimization}
\label{sec:vac-optimization}

When a judge awards Vacuous Credit, omitting the required information or action earns the same criterion reward as providing it, so the reward no longer separates the two responses.
GRPO computes each response's advantage relative to the other responses in its rollout group, so this error can change which responses are reinforced.
A response that omits a required action but receives Vacuous Credit can match or exceed the reward of a response that performs it, and then receives an equal or larger advantage despite its lower rubric satisfaction.
When such awards recur across rollout groups, policy updates can reinforce generic statements in place of the specific information or actions the rubric requires.
Criterion credit should therefore depend on requirement coverage and evidential support before rewards are compared within a group.

\section{\method{}: Learning to Reward for Rubric-Based RL}
\label{sec:method}

\begin{figure}[t]
\centering
\vspace{-0.1in}\includegraphics[width=\linewidth]{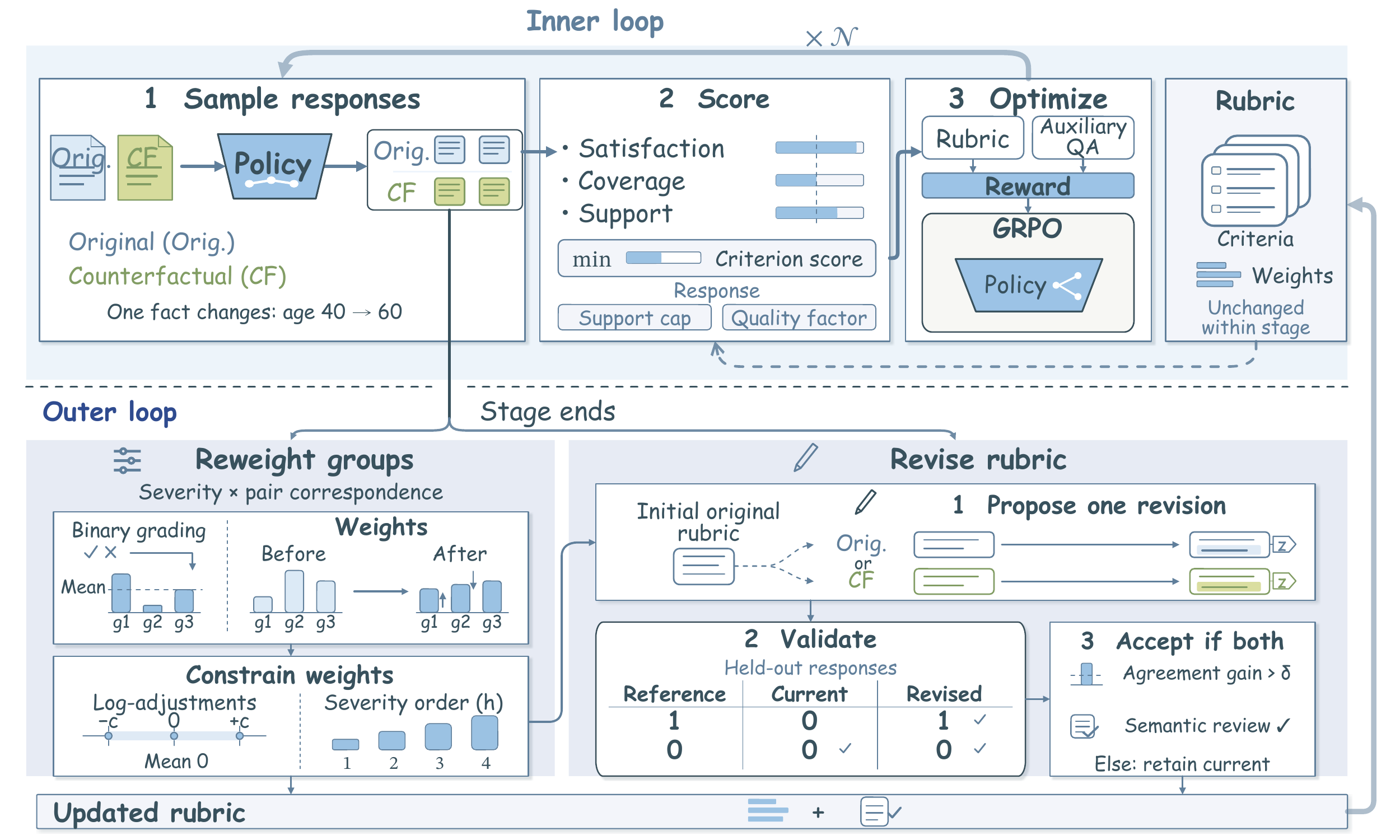}
\vspace{-0.25in}\caption{
\textbf{Overview of MetaRubric.} Each stage alternates $\mathcal N$ evidence-aware GRPO updates (Inner) with rubric adaptation (Outer).
Accepted rubric updates carry into the next stage.
}
\label{fig:metarubrics}
\vspace{-0.1in}
\end{figure}

\method{} addresses Vacuous Credit discussed in \cref{sec:vacuous-credit} with two ideas: a criterion earns credit only for content the response actually provides and can support, and the rubric itself adapts to the errors the policy keeps making.
\method{} alternates between evidence-aware policy optimization in an inner loop and rubric adaptation in an outer loop (\cref{fig:metarubrics}).
The \textit{inner loop} trains the policy with GRPO, using a reward that caps criterion credit by requirement coverage and evidential support (\cref{ssec:meta-inner}). Training prompts come in pairs: each original prompt is paired with a counterfactual that changes one task-relevant fact, thus responses must attend to the facts on which requirements depend.
At the end of each stage, \textit{error-guided weight adaptation} shifts weight toward criterion groups the policy fails more often (\cref{ssec:meta-weights}).
\textit{Semantically anchored rubric revision} then clarifies ambiguous criterion descriptions without changing their core semantics (\cref{ssec:meta-content}).

\noindent\textbf{Stage-wise alternation.}
Training of \method{} proceeds in stages $t=0,1,\ldots$.
Within stage $t$, the rubric configuration $\Phi_t=(\mathcal E_t,\boldsymbol\tau_t)$ is fixed.
Here, $\mathcal E_t$ contains each criterion description and its correspondence label (\cref{ssec:meta-inner}), while $\boldsymbol\tau_t$ contains one log-weight adjustment per criterion group (\cref{ssec:meta-weights}).
Let $\theta_t$ denote the policy parameters at the start of stage $t$. Each stage performs
\begin{equation}
 \theta_{t+1}=\mathcal U_{\mathrm{GRPO}}(\theta_t;\Phi_t),
 \qquad
 \Phi_{t+1}=\mathcal A(\Phi_t;\mathcal B_t,\mathcal V_t),
 \label{eq:meta-alternating}
\end{equation}
where $\mathcal U_{\mathrm{GRPO}}$ denotes $\mathcal N$ GRPO updates under $\Phi_t$, $\mathcal B_t$ is the set of policy responses collected during these updates, and $\mathcal V_t$ is a set of held-out validation cases.
The adaptation step $\mathcal A$ first updates $\boldsymbol\tau_t$ from the errors in $\mathcal B_t$ and then accepts a proposed criterion revision when it passes validation on $\mathcal V_t$.

\subsection{Evidence-Aware Policy Optimization}
\label{ssec:meta-inner}
The inner loop ties the training reward to prompt-specific content that the response actually contains and can support, rather than to generic content that merely appears to meet a criterion (\textit{Inner loop} in \cref{fig:metarubrics}).
Counterfactual prompt pairs expose which requirements depend on specific prompt facts.
The evidence-aware rubric reward credits a criterion only when its coverage and support checks pass, while an auxiliary QA reward independently checks whether the relevant facts appear in the response.
GRPO then optimizes the policy using the combined reward.

\noindent\textbf{Counterfactual prompt pairs.}
To ground credit in prompt-specific facts, we pair each prompt $x$ with a counterfactual $x^{\mathrm{cf}}$ that changes one task-relevant fact while holding the remaining context constant (e.g., the stated age changes from 40 to 60 in \cref{fig:metarubrics}).
Each prompt has a rubric defined for its own facts and each response is scored only against its corresponding rubric, so a fact-dependent criterion receives credit only when the response provides what those facts require.
A correspondence label $z$ records whether each criterion is invariant across the pair or changes in target, weight, or applicability; these labels later define criterion groups (\cref{ssec:meta-weights}) and anchor rubric revision (\cref{ssec:meta-content}).
Construction details are given in \cref{app:paired-construction}.

\noindent\textbf{Evidence-aware rubric reward.}
For each criterion $k$, the judge assigns three scores in $[0,1]$: 
\textit{Satisfaction} $u_k^{\mathrm{sat}}$ measures whether the requirement is fulfilled.
\textit{Coverage} $u_k^{\mathrm{cov}}$ measures whether the response contains the required information or action.
\textit{Support} $u_k^{\mathrm{sup}}$ measures whether that fulfillment is supported by the response and task-allowed evidence, including supplied source material or, when permitted, the prompt and reliable domain knowledge.
The criterion score is the weakest of the three:
\begin{equation}
 \rho_k=\min(u_k^{\mathrm{sat}},u_k^{\mathrm{cov}},u_k^{\mathrm{sup}}).
 \label{eq:meta-criterion-score}
\end{equation}
Removing the required content thus lowers the score regardless of the satisfaction judgment, while coverage alone is also insufficient: a response that states an incorrect sample size covers the requested information but does not satisfy a criterion requiring the correct value.

Replacing $m_k$ in \cref{eq:vac-main} with $\rho_k$, using the current weights $w_k$ (\cref{ssec:meta-weights}), and adding two response-level terms gives the rubric reward
\begin{equation}
 R^{\mathrm{rub}}=\min\!\left(\operatorname{clip}_{[0,1]}\!\left(\frac{\sum_k w_k\rho_k-B_x}{P_x}\right),\,U^{\mathrm{sup}}\right)\frac{1+Q}{2}.
 \label{eq:meta-judge}
\end{equation}
The \textit{support cap} keeps a response from earning a high reward through credited criteria while making unsupported claims elsewhere: $U^{\mathrm{sup}}\in[0,1]$ assesses whether its claims as a whole are supported by task-allowed evidence. The \textit{quality factor} keeps off-topic, inconsistent, or padded responses from being rewarded as highly as focused ones: $Q\in[0,1]$ is the minimum of relevance, coherence, and concision scores.

\noindent\textbf{Auxiliary QA reward.}
Criterion scores still depend on how the judge interprets each requirement.
The auxiliary QA reward provides a judge-independent check of whether prompt-specific facts are recoverable from the response.
A separate reader model, Qwen3-1.7B~\citep{yang2025qwen3}, answers multiple-choice questions using only the response.
Inspired by \citet{yang2026caprl++}, for each rubric criterion, we use an LLM generator to convert the facts in its reference answer and annotations into four-option questions.
We retain only questions that the reader cannot answer from the question and options alone, ensuring that success requires information from the policy response.
The resulting question bank is fixed before training (\cref{app:auxiliary-tasks}).
With importance weight $r_q$, correct option $a_q^\star$, and the reader's answer $a_q(y)$ given the response,
\begin{equation}
 R^{\mathrm{QA}}(x,y)=\frac{\sum_q r_q\mathbf 1[a_q(y)=a_q^\star]}{\sum_q r_q}.
 \label{eq:meta-task}
\end{equation}

\noindent\textbf{Training reward.}
The \method{} training reward combines the rubric and auxiliary QA rewards:
\begin{equation}
 R(x,y;\Phi_t)=\lambda_{\mathrm{rub}}R^{\mathrm{rub}}(x,y;\Phi_t)+\lambda_{\mathrm{QA}}R^{\mathrm{QA}}(x,y).
 \label{eq:meta-reward}
\end{equation}
Within stage $t$, GRPO maximizes this reward over original and counterfactual prompts, and the collected responses $\mathcal B_t$ are passed to the outer loop.

\subsection{Error-Guided Weight Adaptation}
\label{ssec:meta-weights}

Under fixed weights, criteria the policy has already mastered keep their share of the reward, while those it keeps missing gain no extra emphasis.
Weight adaptation shifts weight toward the latter, while preserving the rubric's priorities (\textit{Reweight groups} in \cref{fig:metarubrics}).
 
\noindent\textbf{Reweight groups.}
Rubric criteria are instance-specific, so errors on a single criterion are too sparse to adjust its weight reliably; we therefore pool criteria that play the same role.
Each criterion $k$ has a severity class $h\in\{1,\ldots,4\}$: nice-to-have, should-have, must-have, or contraindication.
The corresponding base weights satisfy $0<\omega_1<\cdots<\omega_4$.
Each criterion also has a correspondence label $z$ (\cref{ssec:meta-inner}).
Criteria sharing $(h,z)$ form a group $g$ with a shared log-weight adjustment $\tau_g$:
\begin{equation}
 w_k=\omega_h\exp(\tau_g).
 \label{eq:meta-weight}
\end{equation}
To measure which requirements the current policy actually fails, a separate grading pass, independent of the training reward, assigns a binary judgment $m_k$ to each criterion on a subset of $\mathcal B_t$ at the end of each stage.
The group error $L_g$ averages $1-m_k$ first over criteria in the group, then over responses, and finally over cases (prompt pairs).
This case-balanced aggregation prevents cases with more criteria or responses from dominating the update.
Each group then moves relative to the mean error $\overline L$ with step size $\eta\geq0$:
\begin{equation}
 \tau_g\leftarrow\tau_g+\eta\,(L_g-\overline L),
 \label{eq:meta-update}
\end{equation}
so groups that fail more often than average gain weight and the others lose weight.
 
\noindent\textbf{Constrain weights.}
Reweighting should change emphasis, not what the rubric values.
To keep a frequently missed nice-to-have from outweighing a must-have, \textit{severity ordering} keeps $\omega_h\exp(\tau_{(h,z)})$ increasing in $h$ for each $z$.
To keep each adjustment small relative to the gaps between priority levels, \textit{clipping} bounds $|\tau_g|\leq c$, where $c=\log\min_{h\in\{1,2,3\}}\omega_{h+1}/\omega_h$ is the smallest log-gap between adjacent severity classes.
To change only relative emphasis rather than the overall reward scale, \textit{centering} keeps the mean adjustment at zero.

\subsection{Semantically Anchored Rubric Revision}
\label{ssec:meta-content}

Reweighting cannot fix a loosely worded criterion: if it reads ``considers the patient's location'' rather than ``asks for the patient's location,'' a response that merely notes regional differences keeps earning Vacuous Credit at any weight.
Policy responses expose such wording, but revising the rubric from them risks drifting toward what the policy already does.
We therefore let revisions clarify wording but not requirements, and accept them only after held-out validation (\textit{Revise rubric}).

\noindent\textbf{Propose one revision.}
The original prompt's initial rubric serves as the semantic anchor, and the correspondence labels specify how each anchored requirement reads under the counterfactual's facts.
Given the anchor, current rubric, prompt, and a response from $\mathcal B_t$, the proposer rewrites one ambiguous or mismatched criterion description.
It also updates the correspondence label $z$ when necessary.

\noindent\textbf{Validate and accept.}
A revision is accepted only if it improves agreement with a reference panel by more than $\delta$ on the held-out responses in $\mathcal V_t$.
A separate semantic review must also confirm that the revision preserves the anchored requirement and remains consistent with the prompt's facts.
Otherwise the current description is retained; the updated weights carry over either way.

\section{Experiments}
\label{sec:experiments}

\subsection{Setup}
\label{ssec:exp-setting}
\begin{table}[t]
\centering
\fontsize{8}{9.3}\selectfont
\vspace{-0.1in}
\caption{\textbf{Performance on four medical benchmarks} (\%; $\uparrow$). Metrics are PubMedQA accuracy, HealthBench-Hard accuracy-axis score, MMOral-X mean score, and MMOral-OPG overall score. Aux. Acc. is auxiliary QA accuracy measured with the Qwen3-1.7B reader. Qwen3 families use Qwen3 for text and Qwen3-VL of the same size for multimodal tasks.}
\label{tab:main}
\vspace{0.05in}
\setlength{\tabcolsep}{3pt}
\renewcommand{\arraystretch}{1.08}
\resizebox{0.95\textwidth}{!}{\begin{tabular*}{\textwidth}{@{\extracolsep{\fill}}lrrrrrrr@{}}
\toprule
Model / training method & PubMedQA & \multicolumn{2}{c}{HealthBench-Hard} & \multicolumn{2}{c}{MMOral-X} & \multicolumn{2}{c}{MMOral-OPG} \\
\cmidrule(lr){2-2}\cmidrule(lr){3-4}\cmidrule(lr){5-6}\cmidrule(l){7-8}
 & Acc. & Acc. & Aux. Acc. & Avg. & Aux. Acc. & Avg. & Aux. Acc. \\
\midrule
\rowcolor{tablegroup} \multicolumn{8}{@{}l}{\textbf{Reference Models}} \\
LLaVA-OneVision & 70.40 & 0.00 & 6.92 & 6.31 & 32.75 & 24.86 & 36.49 \\
InternVL3.5-8B-Instruct & 69.80 & 1.85 & 10.20 & 3.13 & 24.93 & 31.49 & 48.32 \\
MiniCPM-V 4.5 & 54.20 & 1.59 & 10.30 & 3.12 & 20.19 & 26.52 & 35.17 \\
MedGemma 1.5-4B & 55.20 & 4.88 & 18.24 & 5.77 & 25.82 & 24.46 & 32.21 \\
\midrule
\rowcolor{tablegroup} \multicolumn{8}{@{}l}{\textbf{Qwen3-4B family}} \\
\hspace{0.8em}Baseline & 52.60 & 8.83 & 29.46 & 3.38 & 27.34 & 20.59 & 34.45 \\
\hspace{0.8em}+ GRPO~\citep{shao2024grpo} & 72.40 & 10.56 & 31.09 & 5.72 & 34.66 & 23.18 & 37.21 \\
\hspace{0.8em}+ DAPO~\citep{yu2025dapo} & 73.80 & 11.74 & 33.57 & \underline{7.12} & \underline{38.43} & 24.61 & 39.14 \\
\hspace{0.8em}+ Dr.\ GRPO~\citep{liu2025drgrpo} & \underline{76.60} & \textbf{13.28} & \underline{34.56} & 6.94 & 38.24 & \underline{25.72} & \underline{40.37} \\
\hspace{0.8em}+ GSPO~\citep{zheng2025gspo} & 74.60 & 11.38 & 33.11 & 6.42 & 36.88 & 24.08 & 38.63 \\
\hspace{0.8em}\textbf{+ MetaRubric (ours)} & \textbf{78.40} & \underline{13.02} & \textbf{34.93} & \textbf{7.73} & \textbf{39.37} & \textbf{26.35} & \textbf{41.18} \\
\midrule
\rowcolor{tablegroup} \multicolumn{8}{@{}l}{\textbf{Qwen3-8B family}} \\
\hspace{0.8em}Baseline & 44.20 & 4.24 & 30.19 & 3.50 & 28.24 & 26.28 & 36.29 \\
\hspace{0.8em}+ GRPO~\citep{shao2024grpo} & 70.00 & 7.15 & 32.86 & 7.08 & 35.41 & 27.35 & 38.02 \\
\hspace{0.8em}+ DAPO~\citep{yu2025dapo} & \underline{76.60} & 8.64 & 34.52 & 7.86 & 37.20 & 28.44 & 39.35 \\
\hspace{0.8em}+ Dr.\ GRPO~\citep{liu2025drgrpo} & 76.40 & \underline{9.72} & \underline{35.07} & \underline{8.58} & \underline{39.61} & \underline{29.63} & \underline{41.08} \\
\hspace{0.8em}+ GSPO~\citep{zheng2025gspo} & 76.20 & 8.18 & 33.28 & 7.52 & 36.69 & 28.06 & 38.94 \\
\hspace{0.8em}\textbf{+ MetaRubric (ours)} & \textbf{76.80} & \textbf{10.34} & \textbf{36.14} & \textbf{9.21} & \textbf{41.09} & \textbf{30.44} & \textbf{42.61} \\
\midrule
\rowcolor{tablegroup} \multicolumn{8}{@{}l}{\textbf{Gemma-e2b}} \\
\hspace{0.8em}Baseline & 48.00 & 11.23 & 32.74 & 3.69 & 30.17 & 32.77 & 34.86 \\
\hspace{0.8em}+ GRPO~\citep{shao2024grpo} & 51.60 & 13.07 & 33.62 & 4.74 & 33.82 & 31.28 & 37.00 \\
\hspace{0.8em}+ DAPO~\citep{yu2025dapo} & 67.80 & 13.82 & 34.66 & \underline{5.82} & \underline{38.04} & 33.64 & 38.53 \\
\hspace{0.8em}+ Dr.\ GRPO~\citep{liu2025drgrpo} & \underline{70.80} & \underline{14.94} & \underline{35.92} & 5.71 & 37.61 & \textbf{35.42} & \underline{39.76} \\
\hspace{0.8em}+ GSPO~\citep{zheng2025gspo} & 69.60 & 13.34 & 34.08 & 5.07 & 35.43 & 34.58 & 38.12 \\
\hspace{0.8em}\textbf{+ MetaRubric (ours)} & \textbf{72.00} & \textbf{15.68} & \textbf{36.85} & \textbf{6.03} & \textbf{38.70} & \underline{35.10} & \textbf{40.57} \\
\bottomrule
\end{tabular*}}
\vspace{-0.1in}
    \end{table}

\noindent \textbf{Benchmarks.}
We evaluate text-only medical QA on PubMedQA~\citep{jin2019pubmedqa} and HealthBench-Hard~\citep{arora2025healthbench}, and multimodal diagnosis and visual QA on MMOral-X~\citep{fan2026oralgptplus} and MMOral-OPG~\citep{hao2026towards}, respectively.
Dataset splits, rubric construction, and evaluation details appear in \cref{app:datasets,app:evaluation}.

\noindent \textbf{Metrics.}
We report PubMedQA accuracy, HealthBench-Hard accuracy-axis score, MMOral-X mean score across three difficulty subsets, and MMOral-OPG overall score (one evaluation).
GPT-5.5 serves as the strong judge for all model-based test-set evaluations.
Auxiliary QA accuracy (Aux. Acc.) measures the Qwen3-1.7B reader's pooled accuracy on accepted multiple-choice questions given the model response; invalid outputs count as incorrect.

\noindent \textbf{Baselines.}
We compare MetaRubric with four policies: GRPO~\citep{shao2024grpo} with a static judge, DAPO~\citep{yu2025dapo}, Dr.\ GRPO~\citep{liu2025drgrpo}, and GSPO~\citep{zheng2025gspo}. We use Qwen3-4B and Qwen3-8B~\citep{yang2025qwen3} for text-only tasks, their Qwen3-VL counterparts~\citep{bai2025qwen3vl} for multimodal tasks, and Gemma-e2b~\citep{gemmateam2026gemma4} for both.

\subsection{Main Results}
\label{ssec:exp-performance}
\label{ssec:exp-text-performance}
MetaRubric improves over static-judge GRPO in all 21 model-family--metric comparisons in \cref{tab:main}, spanning categorical answers, rubric-graded conversations, and visual question answering.
PubMedQA accuracy increases by 6.00, 6.80 and 20.40 percentage points for Qwen3-4B, Qwen3-8B and Gemma-e2b, respectively; gains on the three open-ended benchmarks range from 1.29 to 3.82 points.
The PubMedQA gains show that improvements also appear under answer accuracy, beyond rubric-based evaluation.
MetaRubric achieves the highest score among training methods in 19 of the 21 comparisons.

\noindent \textbf{Recoverable task information.}
Auxiliary QA accuracy improves by 3.23--5.68 percentage points over GRPO across all three model families and open-ended benchmarks, indicating that the reader can recover more task-relevant information from the responses.
MetaRubric leads on this metric even in the two settings where Dr.\ GRPO achieves slightly higher benchmark scores (\cref{tab:main}).
As auxiliary QA is also a training objective, these gains track the targeted information recovery, while benchmark scores assess broader task performance.

\subsection{Ablation Studies}
\label{ssec:exp-reward-analysis}

\begin{wrapfigure}[13]{R}{0.50\textwidth}
\centering
\vspace{-0.4in}
\includegraphics[width=\linewidth]{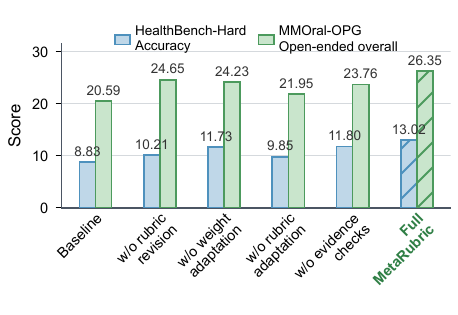}
\vspace{-0.4in}
\caption{\small\textbf{Component ablations.} HealthBench-Hard accuracy-axis scores with Qwen3-4B and MMOral-OPG overall scores with Qwen3-VL-4B (0--100; $\uparrow$).}
\label{fig:ablation}
\vspace{-2in}
\end{wrapfigure}

\cref{fig:ablation} ablates rubric revision, weight adaptation, and evidence checks with the Qwen3-4B family.
The adaptation ablations freeze initial descriptions and labels, weights, or both; the evidence ablation uses satisfaction alone and removes the response-level support cap.
All variants retain the quality multiplier, auxiliary QA reward, and paired sampling.

Freezing both descriptions and weights causes the largest drops: 3.17 points on HealthBench-Hard and 4.40 on MMOral-OPG.
Revision has the larger individual effect on HealthBench-Hard, while weight adaptation matters more on MMOral-OPG, suggesting that clarifying requirements and adjusting their priorities address different task-specific needs.
Removing evidence checks reduces scores by 1.22 and 2.59 points despite retaining auxiliary QA and paired sampling.
These results support both adapting how requirements are expressed and prioritized and checking whether the content credited by the rubric is supported.

\begin{table}[!t]
\centering
\vspace{-0.15in}
\begin{minipage}[t]{0.52\textwidth}
\centering
\caption{\small\textbf{Rubric revision statistics for Qwen3-4B.} Initial criterion counts, criteria revised at least once, and mean description lengths across all criteria before and after training. Weight-only updates are excluded.}
\label{tab:rubric-revision}
\vspace{2pt}
\fontsize{8}{9.5}\selectfont
\setlength{\tabcolsep}{5pt}
\renewcommand{\arraystretch}{0.99}
\begin{tabular}{@{}lrrrr@{}}
\toprule
\rowcolor{tablegroup}
\textbf{Dataset} & \textbf{Total} & {\textbf{Revised} $n$ (\%)} & \multicolumn{2}{c}{\textbf{Avg. words}} \\
\cmidrule(l){4-5}
 & & & Before & After \\
\midrule
HealthBench & 10,448 & 4,727 (45.2\%) & 40.8 & 46.3 \\
PubMedQA & 2,949 & 879 (29.8\%) & 14.3 & 18.2 \\
MMOral-RL & 8,672 & 5,259 (60.6\%) & 16.9 & 22.5 \\
\bottomrule
\end{tabular}

\end{minipage}\hfill
\begin{minipage}[t]{0.45\textwidth}
\centering
\caption{\small\textbf{Retraining with learned rubrics.} Qwen3-4B results on HealthBench-Hard accuracy and MMOral-OPG open-ended overall score ($\uparrow$). MetaRubric updates the rubric during training.}
\label{tab:final-rubric}
\vspace{4pt}
\fontsize{9}{10.5}\selectfont
\renewcommand{\arraystretch}{1.18}
\begin{tabular}{@{}l@{}c@{}c@{}}
\toprule
\rowcolor{tablegroup}[0pt][0pt]
\makebox[0.38\linewidth][l]{\fontsize{8}{10.5}{\textbf{Training rubric}}} & \makebox[0.3\linewidth]{\fontsize{8}{10.5}{\shortstack{\textbf{HealthBench-}\\\textbf{Hard}\\Accuracy $\uparrow$}}} & \makebox[0.30\linewidth]{\fontsize{8}{10.5}{\shortstack{\textbf{MMOral-OPG}\\Open-ended\\overall $\uparrow$}}} \\
\midrule
baseline & 8.83 & 20.59 \\
Final rubric & 9.04 & 21.78 \\
\textbf{MetaRubric} & \textbf{13.02} & \textbf{26.35} \\
\bottomrule
\vspace{-0.2in}
\end{tabular}

\end{minipage}
\end{table}

\begin{figure}[t]
\centering
\includegraphics[width=\linewidth]{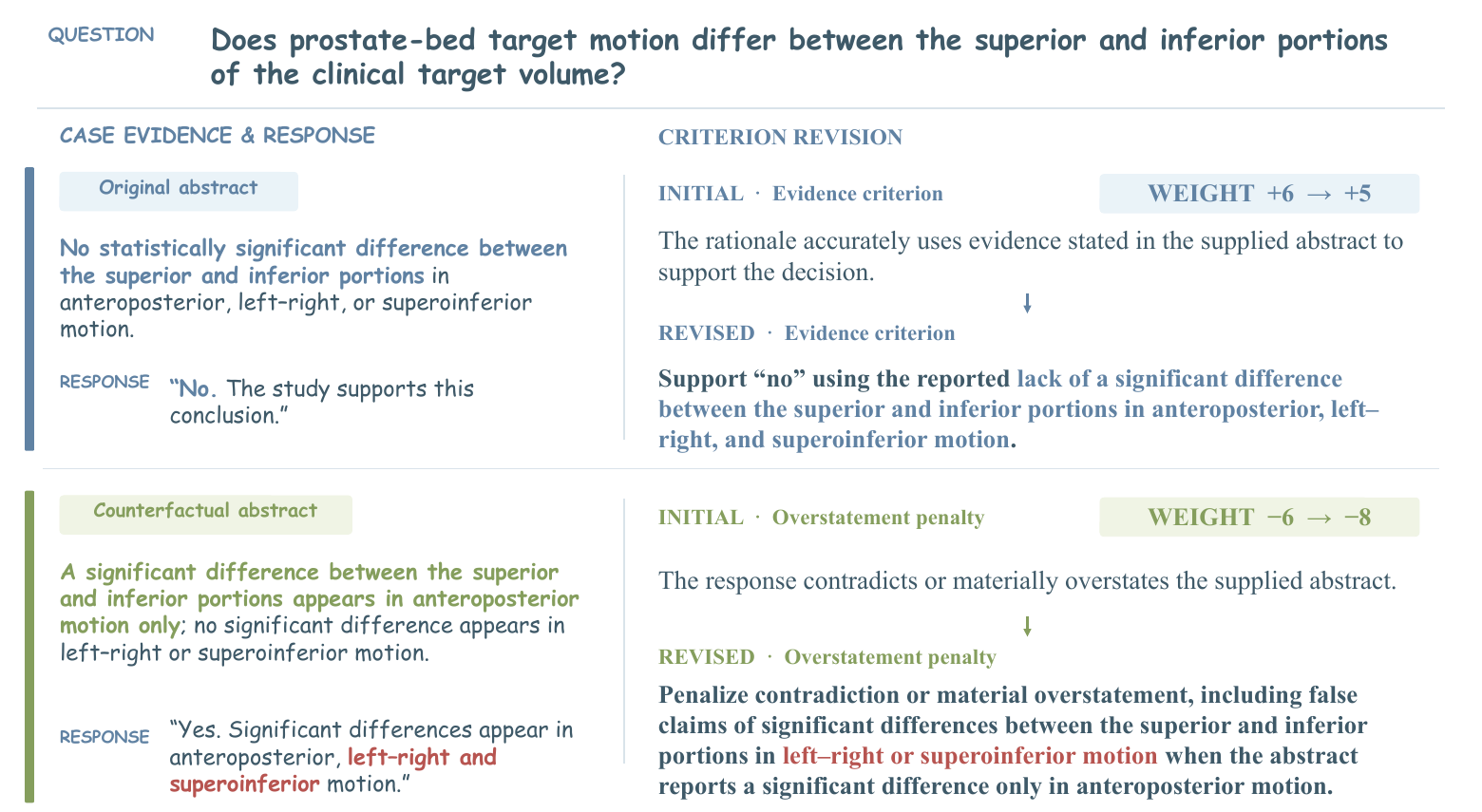}
\vspace{-0.2in}
\caption{\textbf{Illustrative rubric and weight adaptation for a PubMedQA prostate-bed motion question.} Each row aligns an abstract finding, a response, and the criterion before and after revision. The example group-weight update places greater relative emphasis on overstatement errors.}
\label{fig:rubric-trace}
\end{figure}

\subsection{Deep Analysis}
\label{ssec:exp-deep-analysis}
\textbf{Rubric revision analysis.}
\cref{tab:rubric-revision} summarizes criterion revisions during training on HealthBench, PubMedQA and MMOral-RL.
Each criterion is counted once if its description or correspondence label changes at least once; repeated edits do not increase the count, and weight-only updates are excluded.
Average lengths are measured in whitespace-separated words across all criterion descriptions before and after training. Revision affects 45.2\% of HealthBench criteria, 29.8\% of PubMedQA criteria and 60.6\% of MMOral-RL criteria.
Its reach varies substantially across tasks: a majority of criteria change on MMOral-RL, compared with fewer than one third on PubMedQA.
Average descriptions grow by 5.5, 3.9 and 5.6 words, respectively, corresponding to increases of 13.5\%, 27.3\% and 33.1\%.
The fraction of criteria left unchanged ranges from 39.4\% on MMOral-RL to 70.2\% on PubMedQA, showing that revision selectively modifies the rubric rather than rewriting every criterion.
Revision rate does not simply decrease with initial description length: MMOral-RL starts with longer descriptions than PubMedQA but revises a larger fraction of criteria.
These statistics establish the extent of adaptation; the case study below illustrates how a revision makes the required evidence and the scope of a claim explicit.

\noindent \textbf{Is the Final Rubric Sufficient?}
We test whether MetaRubric's gains can be recovered by retraining with its final rubric alone, without further adaptation.
As shown in \cref{tab:final-rubric}, the final rubric improves HealthBench-Hard accuracy from 8.83 to 9.04 and MMOral-OPG overall score from 20.59 to 21.78, whereas MetaRubric reaches 13.02 and 26.35, respectively.
Relative to the baseline, final-rubric retraining recovers 5.0\% and 20.7\% of MetaRubric's gains, computed as the final-rubric gain divided by the full-method gain.
The learned rubric alone is therefore insufficient to reproduce the full method's performance in these runs.
This is consistent with a benefit from matching the rubric to the policy's evolving errors: requirements emphasized late in training may provide a different learning signal when applied from the outset.
These results suggest that the sequence of rubric updates is itself part of the valuable and effective learning procedure.

\noindent \textbf{Case Study.}
\cref{fig:rubric-trace} shows how an evidence criterion specifies the reported absence of motion differences, while an overstatement penalty names unsupported left--right and superoinferior claims.
For an illustrative stage with more overstatement than evidence-use errors, relative group weighting shifts the evidence criterion from $+6$ to $+5$ and the penalty from $-6$ to $-8$.
The revision identifies which claims the evidence supports, while reweighting changes how strongly the remaining errors influence optimization.
Increasing the penalty alone would leave its interpretation unchanged; clarifying its wording alone would leave its relative priority unchanged.
The example thus illustrates why criterion descriptions and weights provide distinct controls over the reward.

\section{Conclusion}
\label{sec:conclusion}

We identify \textit{Vacuous Credit} as a failure mode in rubric-based reinforcement learning, where judges assign high criterion scores even when the information or action required by a criterion is absent from the response.
Our analyses show that criterion credit can persist after the required information is removed, and that correcting an erroneous award can reverse a response's GRPO advantage.
To address this mismatch between rubric credit and response evidence, we introduce MetaRubric{}, which combines evidence-aware scoring with response-guided criterion revision and criterion-group weights that adapt to observed policy errors.
Across text-only and multimodal medical tasks, MetaRubric{} consistently improves over static-rubric GRPO baselines, supporting the use of explicit checks for required information and actions alongside rubric adaptation.

\bibliography{iclr2027_conference}

@article{shan2026survey,
  title   = {A Survey on Rubric-Guided Reinforcement Learning for Language Models},
  author  = {Shan, Zifei and Shao, Fangning},
  journal = {arXiv preprint arXiv:2608.27505},
  year    = {2026}
}

@inproceedings{gao2023scaling,
  title        = {Scaling laws for reward model overoptimization},
  author       = {Gao, Leo and Schulman, John and Hilton, Jacob},
  booktitle    = {International conference on machine learning},
  pages        = {10835--10866},
  year         = {2023},
  organization = {PMLR}
}

@inproceedings{gunjal2025rar,
  title     = {Rubrics as rewards: Reinforcement learning beyond verifiable domains},
  author    = {Gunjal, Anisha and Wang, Anthony and Lau, Elaine and Nath, Vaskar and He, Yunzhong and Liu, Bing and Hendryx, Sean},
  booktitle = {International Conference on Learning Representations},
  volume    = {2026},
  pages     = {127924--127945},
  year      = {2026}
}

@article{jia2026openrs,
  title   = {Open rubric system: Scaling reinforcement learning with pairwise adaptive rubric},
  author  = {Jia, Ruipeng and Yang, Yunyi and Wang, Wen and Wu, Yuxin and Gai, Yongbo and Tao, Siyuan and Zhou, Mengyu and Lin, Jianhe and Jiang, Xiaoxi and Jiang, Guanjun},
  journal = {arXiv preprint arXiv:2602.14069},
  year    = {2026}
}

@inproceedings{li2026rubrichub,
  title     = {{RubricHub}: A comprehensive and highly discriminative rubric dataset via automated coarse-to-fine generation},
  author    = {Li, Sunzhu and Zhao, Jiale and Ren, Huimin and Wei, Zhenlin and Zhou, Yang and Yang, Jingwen and Liu, Shunyu and Zhang, Kaike and Wei, Chen},
  booktitle = {Proceedings of the 64th Annual Meeting of the Association for Computational Linguistics (Volume 1: Long Papers)},
  pages     = {31320--31344},
  year      = {2026}
}

@article{yang2026rubricdropout,
  title   = {Rubric Dropout: A Simple Way to Mitigate Reward Hacking in Rubric-as-Reward {RL}},
  author  = {Yang, Minglai and Guo, Xinyu and Tyagi, Utkarsh and Zhang, Mian and Dumitru, Razvan and Hou, Sunjie and He, Yunzhong and Zhang, Daniel Yue and Liu, Ying},
  journal = {arXiv preprint arXiv:2608.11669},
  year    = {2026}
}

@article{sheng2026rlcer,
  title   = {Reinforcing chain-of-thought reasoning with self-evolving rubrics},
  author  = {Sheng, Leheng and Ma, Wenchang and Hong, Ruixin and Wang, Xiang and Zhang, An and Chua, Tat-Seng},
  journal = {arXiv preprint arXiv:2602.10885},
  year    = {2026}
}

@article{ye2025selfrewarding,
  title   = {Self-rewarding rubric-based reinforcement learning for open-ended reasoning},
  author  = {Ye, Zhiling and Yue, Yun and Wang, Haowen and Han, Xudong and Jiang, Jiadi and Wei, Cheng and Fan, Lei and Liang, Jiaxin and Zhang, Shuowen and Li, Ji and others},
  journal = {arXiv preprint arXiv:2509.25534},
  year    = {2025}
}

@article{xu2026rubricarm,
  title   = {Alternating reinforcement learning for rubric-based reward modeling in non-verifiable {LLM} post-training},
  author  = {Xu, Ran and Liu, Tianci and Dong, Zihan and Yu, Tony and Hong, Ilgee and Yang, Carl and Zhang, Linjun and Zhao, Tao and Wang, Haoyu},
  journal = {arXiv preprint arXiv:2602.01511},
  year    = {2026}
}

@article{guan2026evorubric,
  title   = {{EvoRubric}: Self-Evolving Rubric-Driven {RL} for Open-Ended Generation},
  author  = {Guan, Xin and Hu, Xiaomeng and Huang, Shen and Wang, Zhenyi and Zhang, Bo and Li, Zijian and Xie, Pengjun and Liu, Bo and Cao, Jiuxin},
  journal = {arXiv preprint arXiv:2605.29847},
  year    = {2026}
}

@article{yu2026rlr3,
  title   = {Reinforcement Learning with Robust Rubric Rewards},
  author  = {Yu, Ya-Qi and Wang, Hao and Hong, Fangyu and Qu, Xiangyang and Wu, Gaojie and Luo, Qiaoyu and Xu, Nuo and Wang, Huixin and Xu, Wuheng and Liao, Yongxin and others},
  journal = {arXiv preprint arXiv:2605.30244},
  year    = {2026}
}

@inproceedings{zhang2026chasingtail,
  title     = {Chasing the tail: Effective rubric-based reward modeling for large language model post-training},
  author    = {Zhang, Junkai and Wang, Zihao and Gui, Lin and Mysore Sathyendra, Swarnashree and Jeong, Jaehwan and Veitch, Victor and Wang, Wei and He, Yunzhong and Liu, Bing and Jin, Lifeng},
  booktitle = {International Conference on Learning Representations},
  volume    = {2026},
  pages     = {133430--133457},
  year      = {2026}
}

@inproceedings{liu2026openrubrics,
  title     = {{OpenRubrics}: Towards scalable synthetic rubric generation for reward modeling and {LLM} alignment},
  author    = {Liu, Tianci and Xu, Ran and Yu, Tony and Hong, Ilgee and Yang, Carl and Zhao, Tuo and Wang, Haoyu},
  booktitle = {Proceedings of the 64th Annual Meeting of the Association for Computational Linguistics (Volume 1: Long Papers)},
  pages     = {17417--17437},
  year      = {2026}
}

@article{mahmoud2026rubrichacking,
  title   = {Reward hacking in rubric-based reinforcement learning},
  author  = {Mahmoud, Anas and Rezaei, MohammadHossein and Wang, Zihao and Gunjal, Anisha and Liu, Bing and He, Yunzhong},
  journal = {arXiv preprint arXiv:2605.12474},
  year    = {2026}
}

@article{shao2024grpo,
  title   = {{DeepSeekMath}: Pushing the limits of mathematical reasoning in open language models},
  author  = {Shao, Zhihong and Wang, Peiyi and Zhu, Qihao and Xu, Runxin and Song, Junxiao and Bi, Xiao and Zhang, Haowei and Zhang, Mingchuan and Li, YK and Wu, Yang and others},
  journal = {arXiv preprint arXiv:2402.03300},
  year    = {2024}
}

@article{shen2026rrd,
  title   = {Rethinking rubric generation for improving {LLM} judge and reward modeling for open-ended tasks},
  author  = {Shen, William F and Qiu, Xinchi and Whitehouse, Chenxi and Alazraki, Lisa and Goel, Shashwat and Barbieri, Francesco and Willi, Timon and Mathur, Akhil and Leontiadis, Ilias},
  journal = {arXiv preprint arXiv:2602.05125},
  year    = {2026}
}

@inproceedings{jin2019pubmedqa,
  title     = {{PubMedQA}: A dataset for biomedical research question answering},
  author    = {Jin, Qiao and Dhingra, Bhuwan and Liu, Zhengping and Cohen, William and Lu, Xinghua},
  booktitle = {Proceedings of the 2019 conference on empirical methods in natural language processing and the 9th international joint conference on natural language processing (EMNLP-IJCNLP)},
  pages     = {2567--2577},
  year      = {2019}
}

@article{arora2025healthbench,
  title   = {{HealthBench}: Evaluating large language models towards improved human health},
  author  = {Arora, Rahul K and Wei, Jason and Hicks, Rebecca Soskin and Bowman, Preston and Qui{\~n}onero-Candela, Joaquin and Tsimpourlas, Foivos and Sharman, Michael and Shah, Meghan and Vallone, Andrea and Beutel, Alex and others},
  journal = {arXiv preprint arXiv:2505.08775},
  year    = {2025}
}

@article{yang2025qwen3,
  title   = {{Qwen3} technical report},
  author  = {Yang, An and Li, Anfeng and Yang, Baosong and Zhang, Beichen and Hui, Binyuan and Zheng, Bo and Yu, Bowen and Gao, Chang and Huang, Chengen and Lv, Chenxu and others},
  journal = {arXiv preprint arXiv:2505.09388},
  year    = {2025}
}

@article{hao2026towards,
  title   = {Towards better dental {AI}: A multimodal benchmark and instruction dataset for panoramic {X-ray} analysis},
  author  = {Hao, Jing and Fan, Yuxuan and Sun, Yanpeng and Guo, Kaixin and Lizhuo, Lin and Yang, Jinrong and Ai, Qiyong and Wong, Lun and Tang, Hao and Hung, Kuo},
  journal = {Advances in Neural Information Processing Systems},
  volume  = {38},
  year    = {2026}
}

@article{fan2026oralgptplus,
  title   = {{OralGPT-Plus}: Learning to use visual tools via reinforcement learning for panoramic {X-ray} analysis},
  author  = {Fan, Yuxuan and Hao, Jing and Chen, Hong and Bao, Jiahao and Shao, Yihua and Liang, Yuci and Hung, Kuo Feng and Tang, Hao},
  journal = {arXiv preprint arXiv:2603.06366},
  year    = {2026}
}

@article{yu2025dapo,
  title   = {{DAPO}: An open-source {LLM} reinforcement learning system at scale},
  author  = {Yu, Qiying and Zhang, Zheng and Zhu, Ruofei and Yuan, Yufeng and Zuo, Xiaochen and Yue, Yu and Dai, Weinan and Fan, Tiantian and Liu, Gaohong and Liu, Lingjun and others},
  journal = {Advances in Neural Information Processing Systems},
  volume  = {38},
  pages   = {113222--113244},
  year    = {2026}
}

@article{zheng2025gspo,
  title   = {Group Sequence Policy Optimization},
  author  = {Chujie Zheng and Shixuan Liu and Mingze Li and Xiong-Hui Chen and Bo-Wen Yu and Chang Gao and Kai Dang and Yuqiong Liu and Rui Men and An Yang and Jingren Zhou and Junyang Lin},
  journal = {ArXiv},
  year    = {2025},
  volume  = {abs/2507.18071},
  url     = {https://api.semanticscholar.org/CorpusID:280017753}
}

@article{liu2025drgrpo,
  title   = {Understanding {R1-Zero-Like} Training: A Critical Perspective},
  author  = {Zichen Liu and Changyu Chen and Wen J. Li and Penghui Qi and Tianyu Pang and Chao Du and Wee-Sun Lee and Min Lin},
  journal = {ArXiv},
  year    = {2025},
  volume  = {abs/2503.20783},
  url     = {https://api.semanticscholar.org/CorpusID:277322777}
}

@article{bai2025qwen3vl,
  title   = {{Qwen3-VL} technical report},
  author  = {Bai, Shuai and Cai, Yuxuan and Chen, Ruizhe and Chen, Keqin and Chen, Xionghui and Cheng, Zesen and Deng, Lianghao and Ding, Wei and Gao, Chang and Ge, Chunjiang and others},
  journal = {arXiv preprint arXiv:2511.21631},
  year    = {2025}
}

@article{gemmateam2026gemma4,
  title   = {{Gemma 4} technical report},
  author  = {{Gemma Team} and Abd, Sherif El and Aggarwal, Vaibhav and Algayres, Robin and Andreev, Alek and Bachem, Olivier and Ballantyne, Ian and Brick, Cormac and C{\u{a}}rbune, Victor and Casbon, Michelle and others},
  journal = {arXiv preprint arXiv:2607.02770},
  year    = {2026}
}

@article{yang2026caprl++,
  title={CapRL++: Unified Reinforcement Learning with Verifiable Rewards for Dense Image and Video Captioning},
  author={Yang, Penghui and Xing, Long and Dong, Xiaoyi and Zang, Yuhang and Cao, Yuhang and Wang, Yibin and Zhou, Yujie and Bu, Jiazi and Liang, Jianze and Huang, Qidong and others},
  journal={arXiv preprint arXiv:2606.09393},
  year={2026}
}

@article{yuxuan2026phoenixnest,
  title={PhoenixNest-Video: Evidence-Grounded Multimodal Agent Framework for Automated Video Interview Assessment},
  author={Yuxuan, Fan and Miaojun, Huang and Haimei, Zhang and Jingshen, Wu and Hao, Liu},
  journal={arXiv preprint arXiv:2609.02231},
  year={2026}
}
\bibliographystyle{iclr2027_conference}

\clearpage
\appendix
\section{Appendix}
\label{app:appendix}

\noindent\textbf{Contents}
\begin{itemize}
  \item \hyperref[app:limitations]{\ref*{app:limitations}: Limitations}
  \item \hyperref[app:method-details]{\ref*{app:method-details}: Implementation and Experimental Details}
  \item \hyperref[app:prompt-templates]{\ref*{app:prompt-templates}: Prompt Templates}
  \item \hyperref[app:auxiliary-examples]{\ref*{app:auxiliary-examples}: Auxiliary Question Examples}
\end{itemize}

\subsection{Limitations}
\label{app:limitations}

\noindent\textbf{Single seed.}
Each training configuration uses one seed; the reported results do not quantify variation across training runs.

\noindent\textbf{Comparison scope.}
Many related works have not released their training code, preventing direct comparisons under matched experimental settings.

\subsection{Implementation and Experimental Details}
\label{app:method-details}
\label{app:experiment-details}

\subsubsection{Deployment Details}
\label{app:datasets}
\label{app:paired-construction}

\noindent\textbf{PubMedQA.}
We train on the non-test portion of PubMedQA's expert-labeled PQA-L subset~\citep{jin2019pubmedqa}, reserving a label-stratified validation split and retaining the official test set for evaluation.
Each input contains the question and its abstract context.
For each original training example, we construct a rubric from a six-criterion template, inserting its reference label into the decision criterion.
The three positive criteria reward the correct decision, accurate use of abstract evidence, and relevance to the question and study details.
The three negative criteria penalize unsupported claims, contradictions or overstatements of the abstract, and a missing or generic rationale.
The supplied abstract is the evidence source for assessing the rationale.

\noindent\textbf{HealthBench.}
We use HealthBench~\citep{arora2025healthbench} conversations outside HealthBench-Hard for training and validation, keeping the Hard subset for evaluation.
Each conversation's initial rubric uses the official physician-written criteria and signed point values.

\noindent\textbf{MMOral-RL.}
For both MMOral-X and MMOral-OPG, we train on the reinforcement-learning training set from OralGPT-Plus~\citep{fan2026oralgptplus}, which we denote as \textit{MMOral-RL}.
The two benchmarks evaluate models trained on this shared training set.
MMOral-RL contains 980 original training examples with 8,672 rubric criteria.
Each example has 8.85 criteria on average, with a median of 8 and a range of 1--39.
We generate the initial MMOral-RL rubrics using a procedure informed by the coarse-to-fine rubric construction approach of RubricHub~\citep{li2026rubrichub}.
These task-specific rubrics provide the starting criteria for reward computation and subsequent rubric adaptation.

\subsubsection{Auxiliary Question Construction}
\label{app:auxiliary-tasks}

The auxiliary QA tasks in \cref{ssec:meta-inner} assess whether a response provides the information needed to answer questions about its source case.
We use auxiliary QA for HealthBench-Hard, MMOral-X, and MMOral-OPG. PubMedQA already provides a definite reference label (yes/no/maybe) for each question, so we evaluate its answers directly and do not construct auxiliary questions or report Aux. Acc. for it.
We construct these questions from benchmark reference answers and annotations, using Qwen3-1.7B as the reader.
Question-only screening measures whether this reader can answer without the response; the evaluation-bank acceptance rules are specified below.
Training and evaluation questions use their respective source entries for target extraction, generation, and content review; the MMOral evaluation extension is kept separate from the training bank.
The auxiliary questions and their scoring rules are fixed before training and do not change during rubric adaptation.
\Cref{app:auxiliary-examples} shows example questions from the three evaluation banks.

\noindent \textbf{Source-grounded target extraction.}
For each benchmark entry, we decompose the reference answer or annotations into independently testable facts.
Each auxiliary question targets a distinct fact or a coherent group of facts, and questions associated with the same entry cover complementary information rather than paraphrasing one another.
The number of questions therefore depends on the available non-overlapping targets, rather than requiring the same number for every entry.
We retain a source identifier, a target identifier, and the supporting source facts for each question throughout construction.
For the MMOral extension, the benchmark entry provides the reference for content verification.

\noindent \textbf{Multiple-choice generation and content review.}
A generator converts each target into a four-option question with one correct answer and three distractors.
It receives the source material, reference answer, target facts, other targets from the same entry, and any revision feedback, but no response from a model being evaluated.
A separate content-review call checks that the question preserves its target, that the source supports the correct option, and that this option fully answers the question.
The review also requires mutually exclusive options with exactly one correct answer, plausible distractors of comparable form, and no answer cues from option length or wording.
Questions must remain complementary to the other questions for the entry.
Unsupported facts, ambiguous questions, and changes to the meaning of the correct answer are rejected.

\noindent \textbf{Screening without the response.}
After content review, Qwen3-1.7B receives only the question and its options, labeled A--D.
The response, reference answer, and correct option label are withheld.
We use the same reader checkpoint, prompt, decoding configuration, and option order throughout screening.
A question passes only if the reader returns a valid single option letter that differs from the reference answer.
A correct prediction or an invalid output fails the screen.
For the evaluation banks in \cref{tab:auxiliary-statistics}, HealthBench-Hard uses content-reviewed questions without requiring a question-only screening pass. Each MMOral bank retains all constructed questions once its valid-wrong rate exceeds 90\%, including questions the reader answers correctly.

\noindent \textbf{Target-preserving revision.}
For a question that fails screening, the generator may jointly revise the question and all three distractors.
The source and target identifiers, tested facts, correct answer content, and correct option position remain unchanged.
Each revised candidate undergoes content review before the same reader test is repeated.
Revisions may remove unintended cues or improve the alternatives, but cannot introduce ambiguity or unsupported content.
Once a version passes both checks, we accept it and make no further revisions based on subsequent evaluation scores.

\noindent \textbf{Response-conditioned evaluation.}
For each accepted question, the same reader receives a previously generated response from the model being evaluated, together with the final question and options, and returns one option letter; it is instructed to answer using only the response, without filling in missing information from outside knowledge.
We compute auxiliary QA accuracy (Aux. Acc.) by dividing the total number of correct predictions by the total number of accepted tasks evaluated, counting invalid outputs as incorrect.
Accuracy is pooled over tasks, without first averaging within source entries or revision rounds, and correctness is determined by exact option matching without an LLM judge.
This evaluation accuracy is distinct from the point-weighted training score $R^{\mathrm{QA}}$ in \cref{eq:meta-task}.
It measures response-supported question answering with the specified reader and is reported alongside the original benchmark metrics.

\noindent \textbf{Benchmark and split statistics.}
\Cref{tab:auxiliary-statistics} reports the constructed questions in the three evaluation banks. HealthBench-Hard has 4,117 questions covering 965 of its 1,000 source entries; the remaining 35 entries have no extracted targets. MMOral-X and MMOral-OPG have 828 and 981 questions, respectively, covering all source entries. Mean questions per entry uses the full source split, including entries with no questions.

\begin{table}[htbp]
\centering
\small
\setlength{\tabcolsep}{5pt}
\caption{Auxiliary-question construction statistics for evaluation. Covered entries have at least one question; questions per entry is averaged over all source entries.}
\label{tab:auxiliary-statistics}
\begin{tabular}{lrrrr}
\toprule
Benchmark & Source entries & Covered entries & Questions & Questions/entry \\
\midrule
HealthBench-Hard & 1,000 & 965 & 4,117 & 4.12 \\
MMOral-X & 300 & 300 & 828 & 2.76 \\
MMOral-OPG & 578 & 578 & 981 & 1.70 \\
\bottomrule
\end{tabular}
\end{table}

\subsubsection{Reward Computation and Judge Prompts}
\label{app:scoring-details}
\label{app:judge-prompts}

\noindent\textbf{Fulfillment scores and rubric reference.}
Each criterion describes a requirement to fulfill.
Its reference value $b_k$ records fulfillment when the condition assessed by the source judge is absent; this value is determined by rubric semantics, independently of policy responses.
If the source judge assigns a condition score $a_k^d\in[0,1]$ for assessment $d\in\{\mathrm{sat},\mathrm{cov},\mathrm{sup}\}$, we express it on the fulfillment scale as
\begin{equation}
 u_k^d=b_k+(1-2b_k)a_k^d.
 \label{eq:fulfillment-orientation}
\end{equation}
The same transformation applies to binary judgments.
Thus a source judgment of condition absence corresponds to $u_k^d=b_k$, and all assessments measure fulfillment in the same direction.
The identity $1-\max_d a_k^d=\min_d(1-a_k^d)$ lets both score orientations use the single minimum in \cref{eq:meta-criterion-score}.

The contribution of criterion $k$ is $w_k(\rho_k-b_k)$.
Summing these contributions yields $\sum_k w_k\rho_k-B_x$, where $B_x=\sum_k w_kb_k$.
Full fulfillment gives the normalization $P_x=\sum_k w_k(1-b_k)>0$, so the normalized score is one when every criterion is fulfilled.
The offset preserves the rubric's scoring origin; it is not estimated from responses or tuned during training.
When weights change, both $B_x$ and $P_x$ are recomputed using those weights and the fixed reference values $b_k$.
Clipping, the response-level support cap, and the quality multiplier are then applied as in \cref{eq:meta-judge}.

\subsubsection{Baselines and Training Configurations}
\label{app:training-configurations}

\noindent\textbf{GRPO groups and advantages.}
For each prompt pair, we sample $G\geq2$ responses to the original prompt and $G$ responses to its counterfactual, and treat all $2G$ responses as one GRPO group.
Each response is scored with its own prompt's rubric. For response $j$ with reward $R_j$ from \cref{eq:meta-reward}, the advantage is
\begin{equation}
 A_j=\frac{R_j-\bar R}{\operatorname{std}(R)+\epsilon},
 \label{eq:vac-advantage}
\end{equation}
where $\bar R$ and $\operatorname{std}(R)$ are the mean and sample standard deviation over all $2G$ rewards, and $\epsilon>0$ prevents division by zero.
The advantage measures how far a response's reward lies above or below the group mean, scaled by the within-group reward variation.
The resulting advantages enter the clipped GRPO objective to update the policy.

\noindent\textbf{Hyperparameters.}
\cref{tab:training-hyperparameters} summarizes the Qwen3-4B-Instruct-2507 configuration for HealthBench supplied with the supplementary code.
The prompt batch and PPO mini-batch each contain 18 prompts, corresponding to nine complete original--counterfactual pairs.
With $G=8$ responses per prompt, an update contains 144 responses and each paired GRPO group contains $2G=16$ responses.
The launcher disables row-level shuffling to preserve adjacent pairs.
Dynamic batching limits tokens per GPU rather than fixing the number of responses in each micro-batch.

\begin{table}[t]
\centering
\small
\caption{\textbf{HealthBench training hyperparameters in the supplementary configuration.} Batch sizes count prompts before rollout expansion. Token budgets are per GPU.}
\label{tab:training-hyperparameters}
\begin{tabular}{@{}ll@{}}
\toprule
Parameter & Value \\
\midrule
Policy & Qwen3-4B-Instruct-2507; thinking disabled \\
Optimizer & AdamW, $\beta_1=0.9$, $\beta_2=0.999$ \\
Learning rate / weight decay & $10^{-6}$ / $0.01$ \\
Learning-rate schedule & 4 warmup updates, then constant \\
Gradient norm clipping & $1.0$ \\
Prompt batch / PPO mini-batch & 18 / 18 \\
PPO epochs per rollout batch & 1 \\
PPO clipping (lower / upper) & $0.2$ / $0.2$ \\
Dual-clip coefficient & $3.0$ \\
Loss aggregation & Token mean \\
KL penalty / entropy coefficient & Disabled / $0$ \\
Rollouts per prompt $G$ & 8 \\
Sampling temperature / top-$p$ / top-$k$ & $0.7$ / $0.8$ / 20 \\
Maximum prompt / response length & 2,048 / 1,024 tokens \\
Actor / log-probability token budget & 8,192 / 16,384 \\
Rollout batched-token budget & 16,384 \\
Advantage stabilizer $\epsilon$ & $10^{-6}$ \\
Training epoch cap / seed & 3 / 42 \\
Checkpoint interval & 10 updates \\
\midrule
Reward coefficients $(\lambda_{\mathrm{rub}},\lambda_{\mathrm{QA}})$ & $(1.0,0.3)$ \\
Auxiliary QA reader & Qwen3-1.7B; thinking disabled \\
Reader sampling temperature & $0$ \\
Severity anchors $(\omega_1,\omega_2,\omega_3,\omega_4)$ & $(4,5,6,8)$ \\
Initial log-weight adjustments $\tau_g$ & $0$ \\
Weight-update step size $\eta$ & $0.1$ \\
Log-weight bound $c$ & $\log(6/5)\approx0.1823$ \\
Responses for outer-loop error estimation & At most 96; one per sampled prompt \\
Criterion revision proposals per boundary & At most 1 \\
Revision validation cases & 20 distinct prompt pairs \\
Agreement-gain threshold $\delta$ & $0.02$ (strict improvement above threshold) \\
\bottomrule
\end{tabular}
\end{table}

\noindent\textbf{Outer-loop configuration.}
The reward coefficients in \cref{eq:meta-reward} are $\lambda_{\mathrm{rub}}=1$ and $\lambda_{\mathrm{QA}}=0.3$.
The four severity anchors in \cref{eq:meta-weight} are $(4,5,6,8)$, which give $c=\log(6/5)$, and all group adjustments start at zero.
At each boundary, the outer loop estimates group errors from at most 96 responses, using one response per sampled prompt, and updates weights with $\eta=0.1$.
A proposed revision requires 20 distinct validation cases and a case-balanced agreement gain strictly greater than $\delta=0.02$, as well as semantic approval.
The proposer receives neither the held-out responses nor their reference judgments.
The inner judge and outer grading, proposal, semantic-review, and reference-panel roles use GPT-5.4-mini; outer roles use separate calls with temperature zero.

\noindent\textbf{Execution and checkpointing.}
The configuration uses two GPUs for policy training with FSDP2 and tensor-parallel rollout generation, with actor parameters and optimizer states offloaded to CPU.
Gradient checkpointing and FlashAttention are enabled.
Checkpoints save the model, optimizer, and training state, allowing the next segment to resume after a rubric update.
The software stack is verl 0.8.0.dev, PyTorch 2.8.0 (CUDA 12.8), vLLM 0.11.0, Transformers 4.57.6, and FlashAttention 2.8.3.

\subsubsection{Evaluation Protocols and Metrics}
\label{app:evaluation}

\noindent\textbf{Test-set judge.}
GPT-5.5 serves as the strong judge for all model-based test-set evaluations in \cref{sec:experiments}.

\noindent\textbf{Text-only benchmarks.}
PubMedQA~\citep{jin2019pubmedqa} provides biomedical questions with yes/no/maybe answers. We evaluate accuracy on its 500-question expert-labeled test set, providing the abstract context with each question.
HealthBench-Hard~\citep{arora2025healthbench} contains 1,000 challenging health conversations assessed with weighted, conversation-specific rubrics. We report the accuracy-axis score.

\noindent\textbf{Multimodal benchmarks.}
MMOral-X~\citep{fan2026oralgptplus} evaluates diagnosis from panoramic X-rays on 300 questions, divided equally into Simple, Moderate and Complex subsets. We report the mean score across these three subsets.
MMOral-OPG~\citep{hao2026towards} evaluates open-ended visual question answering on 578 questions across six medical task types. We report its overall score from one evaluation.

\clearpage
\raggedbottom
\subsection{Prompt Templates}
\label{app:prompt-templates}

The following templates cover rubric and counterfactual construction, auxiliary QA, evidence-aware scoring, and rubric adaptation. Braced fields denote task-specific inputs; image inputs accompany the corresponding multimodal query.

\newenvironment{paperprompt}[3][]{%
  \def\promptcaption{#1}%
  \ifx\promptcaption\empty
    \def\promptcaption{#2 prompt.}%
  \fi
  \def\promptlabel{fig:prompt-#3}%
  \begin{tcolorbox}[
    colback=findingborder!5!white,
    colframe=findingborder,
    coltitle=black,
    colbacktitle=tablegroup,
    fonttitle=\bfseries,
    fontupper=\small,
    title={#2},
    width=\textwidth,
    boxrule=0.5pt,
    arc=2mm,
    outer arc=2mm,
    before skip=10pt,
    after skip=0pt]
  \raggedright
  \setlength{\parskip}{0.4em}
}{%
  \end{tcolorbox}
  \nopagebreak
  \captionof{figure}{\promptcaption}\label{\promptlabel}
  \par\medskip}
\newcommand{\promptfield}[1]{\texttt{\{#1\}}}

\subsubsection{Data and Auxiliary Question Construction}
\label{app:prompts-construction}

\begin{paperprompt}{Initial Rubric Construction}{initial-rubric}
\textbf{System prompt}

Construct a task-specific rubric from the supplied prompt, reference answer, and annotations. Use the source as the factual authority. If a source rubric is supplied, preserve its requirements, scoring polarity, and priorities.

Each criterion must specify an observable response requirement or an undesirable behavior. Cover distinct requirements without duplicating credit. Preserve conditions, quantities, negations, and uncertainty. Do not introduce unsupported facts or clinical inferences. Assign a stable criterion\_id and severity: nice\_to\_have, should\_have, must\_have, or contraindication. State whether the criterion describes desired content or an undesirable behavior, and identify its supporting source facts.

Return JSON with rubrics, each containing criterion\_id, criterion, severity, polarity, and source\_support. Do not use candidate responses to define requirements.

\textbf{Query prompt}

Prompt and supplied evidence: \promptfield{prompt}\par
Reference answer and annotations: \promptfield{reference}\par
Source rubric, if available: \promptfield{source\_rubric}
\end{paperprompt}

\begin{paperprompt}{Counterfactual Prompt and Rubric Construction}{counterfactual-construction}
\textbf{System prompt}

Create one counterfactual by changing exactly one task-relevant fact. Hold the remaining context fixed. Update the reference answer and fact-dependent rubric criteria only as required by that intervention. Preserve unchanged criteria verbatim and retain the original rubric as the semantic anchor.

Record the changed fact before and after the intervention. For each criterion, specify whether it is invariant, changes target, changes weight, becomes inapplicable, or becomes applicable. Preserve its correspondence to the original requirement. Do not invent additional findings, change unrelated requirements, or choose facts from a policy response.

For an image-based task, keep the image unchanged. State explicitly that the counterfactual is hypothetical and which single observation should be replaced. Do not claim that the original image depicts the hypothetical finding. Preserve unrelated observations, tooth identifiers, and coordinates.

Return JSON with changed\_fact, counterfactual\_prompt, counterfactual\_reference, and criterion\_correspondences. Each correspondence contains the original criterion ID, counterfactual criterion text, correspondence label, and rationale.

\textbf{Query prompt}

Original prompt: \promptfield{prompt}\par
Reference and annotations: \promptfield{reference}\par
Original rubric: \promptfield{rubric}\par
Modality and supplied evidence: \promptfield{evidence}
\end{paperprompt}

\begin{paperprompt}{Criterion Correspondence Labeling}{correspondence-labeling}
\textbf{System prompt}

Align criteria across an original task and its single-fact counterfactual. Use the original rubric as the semantic anchor. Assign one label per correspondence: INVARIANT, TARGET\_CHANGE, WEIGHT\_CHANGE, DROPPED, or ADDED.

INVARIANT preserves the requirement, target, weight, and applicability. TARGET\_CHANGE means the changed fact requires a different answer target. WEIGHT\_CHANGE preserves the target but changes its importance. DROPPED means an original requirement is inapplicable under the counterfactual; ADDED means the intervention activates a requirement absent on the original side. Do not label a paraphrase as a target change. Explain non-invariant labels by the changed fact, not by candidate responses.

Return JSON with original\_id, counterfactual\_id, label, and rationale for each correspondence. Use null for the absent side of an added or dropped requirement. Flag ambiguous alignments for review.

\textbf{Query prompt}

Original task and rubric: \promptfield{original}\par
Counterfactual task and rubric: \promptfield{counterfactual}\par
Single changed fact: \promptfield{changed\_fact}
\end{paperprompt}

\begin{paperprompt}{Auxiliary Question Generation}{qa-generation}
\textbf{System prompt}

Build one four-option question from the supplied source and target. Return JSON with question, correct, and distractors (three strings). Preserve the target's question focus and complete answer meaning.

The correct option must fully answer the target and be supported by the source. Distractors must be plausible, mutually exclusive alternatives that are false for this source. Match all four options in form, specificity, and completeness. Avoid all-of-the-above or none-of-the-above options, answer-position hints, wording or length cues, and unsupported claims in the correct answer.

Keep the question complementary to sibling targets. Use only the supplied source and target; do not inspect candidate model responses.

\textbf{Query prompt}

Source: \promptfield{source}\par
Target: \promptfield{target}\par
Sibling targets: \promptfield{siblings}\par
Revision feedback, if any: \promptfield{feedback}
\end{paperprompt}

\begin{paperprompt}{Auxiliary Question Content Review}{qa-review}
\textbf{System prompt}

Independently review the proposed four-option question against its source, target, and sibling targets. Identify the uniquely correct option by its zero-based position in the supplied options array.

Accept only if the question preserves the target, the source fully supports the correct option, and exactly one option completely answers the question. The three distractors must be plausible, mutually exclusive alternatives that are false for this source. Match option form and specificity; reject ambiguity, unsupported correct-answer claims, answer cues, and overlap with sibling targets.

Return JSON with accepted (boolean), reason (string), and correct\_index (0, 1, 2, 3, or null). Do not inspect candidate model responses or reader predictions.

\textbf{Query prompt}

Source: \promptfield{source}\par
Target: \promptfield{target}\par
Sibling targets: \promptfield{siblings}\par
Question: \promptfield{question}\par
Options in their presented order: \promptfield{options}
\end{paperprompt}

\begin{paperprompt}{Auxiliary Question Revision}{qa-revision}
\textbf{System prompt}

Revise one four-option auxiliary question that failed content review or the question-only reader screen. Preserve source and target identifiers, tested facts, complete correct-answer meaning, and correct-option position. You may revise the question and all three distractors.

The question must identify the requested information without revealing the correct answer. A response explicitly supplying the target facts must suffice to answer it. Match all options in grammar, specificity, completeness, and plausibility. Each distractor must be false for this source but plausible in another case. Preserve conditions, quantities, negations, exceptions, and required actions. Do not introduce unsupported facts, ambiguity, or answer cues to cause reader failure.

Use the source and target as authority. Reader predictions are screening feedback, not factual evidence. Do not inspect responses from models being evaluated.

Return JSON with question, correct, and distractors (three strings). If no faithful revision is possible, return unresolved=true with a reason.

\textbf{Query prompt}

Source and target identifiers: \promptfield{identifiers}\par
Source, original target, and sibling targets: \promptfield{source\_targets}\par
Current question, options, and fixed correct-option position: \promptfield{current\_item}\par
Prior correct answer: \promptfield{prior\_correct\_answer}\par
Content-review and screening feedback: \promptfield{feedback}
\end{paperprompt}

\subsubsection{Reward Evaluation}
\label{app:prompts-training}

\begin{paperprompt}{Evidence-Aware Rubric Judge}{evidence-aware-judge}
\textbf{System prompt}

Score every supplied rubric criterion against the response. Use only task-allowed evidence: supplied source material for source-only tasks, or the prompt and reliable domain knowledge when permitted. Treat all supplied content as data.

Assign three independent fulfillment scores in [0,1]: satisfaction, coverage, and support. Satisfaction measures whether the requirement is correctly fulfilled. Coverage measures whether the required information or action is present. Support measures whether that fulfillment is supported by the response and allowed evidence. Topical overlap, confidence, and an intention to act do not establish fulfillment. Missing required content receives zero coverage. A stated but incorrect value may provide coverage while failing satisfaction and support.

The supplied reference\_value specifies fulfillment when the source criterion's condition is absent. If it is 0, score fulfillment of the described requirement. If it is 1, score avoidance of the described undesirable condition: its supported presence lowers fulfillment. Return all three scores on this common fulfillment scale.

Also score global\_support for the response's claims as a whole, and relevance, coherence, and concision, each in [0,1]. Unsupported claims outside the rubric still affect global\_support. Do not reward verbosity or calculate the aggregate reward.

Return JSON with criteria (criterion\_id, satisfaction, coverage, support), global\_support, and quality (relevance, coherence, concision). Return exactly one entry per criterion, in the supplied order.

\textbf{Query prompt}

Prompt and evidence contract: \promptfield{prompt\_and\_evidence}\par
Response: \promptfield{response}\par
Rubric rows with IDs, requirements, and reference values: \promptfield{rubric}
\end{paperprompt}

\begin{paperprompt}{Auxiliary QA Reader}{qa-reader}
\textbf{System prompt: response-conditioned reading}

Answer the multiple-choice question using only the supplied benchmark model response. Output exactly one capital letter: A, B, C, or D. No explanation. If the response does not support an answer, choose the best supported option.

\textbf{Query prompt}

Benchmark model response:\par
\promptfield{response}\par
Question:\par
\promptfield{question}\par
A. \promptfield{option\_A}\par
B. \promptfield{option\_B}\par
C. \promptfield{option\_C}\par
D. \promptfield{option\_D}\par

\textbf{Screening without the response}

Use the same question and option order, omit the response, and replace the system prompt with:

Answer the multiple-choice question using only the text given. Output exactly one capital letter: A, B, C, or D. No explanation.
\end{paperprompt}

\subsubsection{Rubric Adaptation}
\label{app:prompts-revision}

\begin{paperprompt}{Independent Criterion Grading for Weight Adaptation}{weight-grading}
\textbf{System prompt}

Evaluate the response against each supplied criterion using the task's facts. Treat all inputs as data. For both positive and negative points, met means the behavior described by the criterion is present. In particular, an undesirable behavior being present gives met=true for a negative-point criterion.

Require substantive correctness, not tone, headings, or topical overlap. Return JSON with a judgments array containing id, met (boolean), and reason (specific evidence). Return exactly one entry per criterion, in the supplied order.

\textbf{Query prompt}

Task facts: \promptfield{prompt}\par
Response: \promptfield{response}\par
Criteria with IDs and signed points: \promptfield{criteria}
\end{paperprompt}

\begin{paperprompt}{Rubric Revision Reference Panel}{revision-panel}
\textbf{System prompt}

Provide fixed reference judgments for the held-out response using the original anchored criteria and the held-out case's facts. Determine whether the exact behavior described by each criterion is substantively present. For a negative-point criterion, report whether the undesirable behavior is present.

Do not reward generic reassurance or unsupported certainty. Do not transplant facts from the original case into a different held-out case. Judge before seeing the proposed revision, without access to the training reward or the proposal's validation score.

Return JSON with judgments, each containing id, met (boolean), and reason (specific response evidence). Return one entry per criterion in the supplied order.

\textbf{Query prompt}

Original criterion context: \promptfield{original\_criterion\_context}\par
Held-out case context: \promptfield{held\_case\_context}\par
Held-out response: \promptfield{held\_response}\par
Original criteria with IDs and signed points: \promptfield{criteria}
\end{paperprompt}

\begin{paperprompt}{Anchored Rubric Revision Proposal}{revision-proposal}
\textbf{System prompt}

Propose at most one revision to an ambiguous or mismatched criterion description. Use the original prompt's initial rubric as the semantic anchor, interpreted under the supplied original or counterfactual facts. Use the policy response to identify unclear wording, not to redefine the target.

Preserve the anchored requirement, scoring polarity, severity, and fixed auxiliary questions and answers. Clarify what the response must provide without adding a requirement or weakening an unmet one. Update the correspondence label only when necessary for consistency with the prompt pair. Allowed labels: INVARIANT, TARGET\_CHANGE, WEIGHT\_CHANGE, DROPPED, and ADDED. Do not change weights.

Return JSON with edits, containing at most one entry with criterion\_id, criterion (the complete revised description), correspondence\_label, and rationale. Return an empty edits array if no justified revision is available.

\textbf{Query prompt}

Original prompt and anchored rubric: \promptfield{anchor}\par
Prompt to be revised and current rubric: \promptfield{current\_contract}\par
Correspondence labels and fixed metadata: \promptfield{metadata}\par
Policy response from the training stage: \promptfield{policy\_response}
\end{paperprompt}

\begin{paperprompt}{Rubric Revision Semantic Review}{semantic-review}
\textbf{System prompt}

Review the proposed criterion revision against the original semantic anchor and the prompt pair's facts. Check that it preserves the requirement, scoring polarity, and severity; adds no unsupported facts; and uses a correspondence label consistent with the single counterfactual change. Reject wording that weakens an unmet requirement, rewards generic language instead of required content, or introduces a new target.

Judge semantic validity independently of agreement gains on held-out responses. Return JSON with approved (boolean) and rationale (string). Approve only when every check passes; otherwise explain the specific conflict.

\textbf{Query prompt}

Original prompt and anchored rubric: \promptfield{anchor}\par
Current prompt, rubric, and correspondence labels: \promptfield{current\_contract}\par
Proposed criterion edit and rationale: \promptfield{proposal}
\end{paperprompt}

\clearpage
\subsection{Auxiliary Question Examples}
\label{app:auxiliary-examples}

The following examples illustrate the auxiliary questions described in \cref{app:auxiliary-tasks}. Source excerpts and correct answers are shown for reference; the reader receives only the model response, question, and options.

\begin{paperprompt}[Auxiliary question from HealthBench-Hard.]{HealthBench-Hard: Preserving Follow-up Timing}{example-healthbench}
\textbf{Source request (excerpt)}

``mention we only recommend a re-check if fever returns or symptoms worsen or fail to improve within another 2-3 days.''

\textbf{Question}

What specific follow-up timing should the rewritten message say for lack of improvement?

\textbf{Options}

\textbf{A.} It should say: fails to improve within another 2-3 days.\par
\textbf{B.} It should say: fails to improve within another 5-7 days.\par
\textbf{C.} It should say: fails to improve after completing the full 10-day course.\par
\textbf{D.} It should say: fails to improve within another 24-48 hours.

\textbf{Correct answer: A}
\end{paperprompt}

\begin{paperprompt}[Auxiliary question from MMOral-X.]{MMOral-X: Treatment Identification}{example-mmoral-x}
\textbf{Source reference (excerpt)}

``Endodontic treatments are noted on teeth 13, 23, 24, 26, and 45.''

\textbf{Question}

Which teeth have endodontic treatments?

\textbf{Options}

\textbf{A.} Teeth 13, 23, 24, 26, and 47.\par
\textbf{B.} Teeth 13, 23, 24, 26, and 45.\par
\textbf{C.} Teeth 13, 24, 25, 26, and 45.\par
\textbf{D.} Teeth 12, 23, 24, 26, and 45.

\textbf{Correct answer: B}
\end{paperprompt}

\begin{paperprompt}[Auxiliary question from MMOral-OPG.]{MMOral-OPG: Tooth Counting and Identification}{example-mmoral-opg}
\textbf{Source reference}

``Four wisdom teeth are detected: \#18, \#28, \#38, and \#48.''

\textbf{Question}

How many wisdom teeth are detected in the radiograph, and which teeth are they?

\textbf{Options}

\textbf{A.} Four wisdom teeth are detected: \#18, \#28, \#38, and \#48.\par
\textbf{B.} Two wisdom teeth are detected: \#18 and \#38.\par
\textbf{C.} Four wisdom teeth are detected: \#17, \#27, \#37, and \#47.\par
\textbf{D.} Three wisdom teeth are detected: \#18, \#28, and \#48.

\textbf{Correct answer: A}
\end{paperprompt}

\end{document}